\documentclass{article}

\usepackage{microtype}
\usepackage{graphicx}
\usepackage{subfigure}
\usepackage{booktabs} 
\usepackage{multirow}
\usepackage{lipsum}
\usepackage{wrapfig}
\usepackage{amsmath}
\usepackage{amsthm}
\usepackage{amssymb}
\usepackage{xcolor}
\usepackage{soul}
\newcommand{\mathcolorbox}[2]{\colorbox{#1}{$\displaystyle #2$}}
\usepackage{hyperref}
\newcommand{\bfx}{\mathbf{x}}
\newcommand{\bfs}{\mathbf{s}}
\newcommand{\bfz}{\mathbf{z}}
\newcommand{\bff}{\mathbf{f}}

\DeclareMathOperator{\TorsionAngle}{TorsionAngle}
\DeclareMathOperator{\BetaCarbons}{BetaCarbons}

\DeclareMathOperator{\RMSDAlign}{RMSDAlign}
\DeclareMathOperator{\HarmonicPrior}{HarmonicPrior}

\DeclareMathOperator{\AlphaFoldEmbedder}{AlphaFoldEmbedder}
\DeclareMathOperator{\InputEmbedder}{InputEmbedder}
\DeclareMathOperator{\Evoformer}{Evoformer}
\DeclareMathOperator{\StructureModule}{StructureModule}
\newcommand{\pluseq}{\mathrel{+}=}
\DeclareMathOperator{\Bin}{Bin}

\DeclareMathOperator{\Linear}{Linear}

\DeclareMathOperator{\GaussianFourierEmbedding}{GaussianFourierEmbedding}
\DeclareMathOperator{\length}{length}
\DeclareMathOperator{\OneHot}{OneHot}
\DeclareMathOperator{\Concat}{concat}

\DeclareMathOperator{\TriangleAttentionStartingNode}{TriangleAttentionStartingNode}
\DeclareMathOperator{\TriangleAttentionEndingNode}{TriangleAttentionEndingNode}
\DeclareMathOperator{\TriangleMultiplicationOutgoing}{TriangleMultiplicationOutgoing}
\DeclareMathOperator{\TriangleMultiplicationIncoming}{TriangleMultiplicationIncoming}
\DeclareMathOperator{\PairTransition}{PairTransition}
\usepackage{multirow,makecell}

\usepackage[accepted]{icml2024}

\usepackage{amsmath}
\usepackage{amssymb}
\usepackage{mathtools}
\usepackage{amsthm}

\usepackage[capitalize,noabbrev]{cleveref}

\theoremstyle{plain}

\theoremstyle{definition}

\theoremstyle{remark}

\usepackage[textsize=tiny]{todonotes}

\icmltitlerunning{Improving Alpha\textsc{Flow} for Efficient Protein Ensembles Generation}

\begin{document}

\twocolumn[
\icmltitle{Improving Alpha\textsc{Flow} for Efficient Protein Ensembles Generation}



\icmlsetsymbol{equal}{*}

\begin{icmlauthorlist}
\icmlauthor{Shaoning Li}{equal,1}
\icmlauthor{Mingyu Li}{equal,2}
\icmlauthor{Yusong Wang}{3}
\icmlauthor{Xinheng He}{4}
\icmlauthor{Nanning Zheng}{3}
\icmlauthor{Jian Zhang}{2}
\icmlauthor{Pheng Ann Heng}{1}
\end{icmlauthorlist}

\icmlaffiliation{1}{Department of Computer Science and Engineering, The Chinese University of Hong Kong}
\icmlaffiliation{2}{Medicinal Chemistry and Bioinformatics Center, Shanghai Jiao Tong University School of Medicine}
\icmlaffiliation{3}{National Key Laboratory of Human-Machine Hybrid Augmented Intelligence
and Institute of Artificial Intelligence and Robotics, Xi'an Jiaotong University}
\icmlaffiliation{4}{Shanghai Institute of Materia Medica, Chinese Academy of Sciences}

\icmlcorrespondingauthor{Pheng Ann Heng}{pheng@cse.cuhk.edu.hk}
\icmlcorrespondingauthor{Jian Zhang}{jian.zhang@sjtu.edu.cn}

\icmlkeywords{Machine Learning, ICML}

\vskip 0.3in
]



\printAffiliationsAndNotice{\icmlEqualContribution} 

\begin{abstract}
Investigating conformational landscapes of proteins is a crucial way to understand their biological functions and properties.
Alpha\textsc{Flow} stands out as a \textit{sequence-conditioned} generative model that introduces flexibility into structure prediction models by fine-tuning AlphaFold under the flow-matching framework.
Despite the advantages of efficient sampling afforded by flow-matching, Alpha\textsc{Flow} still requires multiple runs of AlphaFold to finally generate one single conformation.
Due to the heavy consumption of AlphaFold, its applicability is limited in sampling larger set of protein ensembles or the longer chains within a constrained timeframe.
In this work, we propose a \textit{feature-conditioned} generative model called \underline{Alpha\textsc{Flow}-Lit} to realize efficient protein ensembles generation.
In contrast to the full fine-tuning on the entire structure, we focus solely on the light-weight structure module to reconstruct the conformation.
Alpha\textsc{Flow}-Lit performs on-par with Alpha\textsc{Flow} and surpasses its distilled version without pretraining, all while achieving a significant sampling acceleration of around 47 times.
The advancement in efficiency showcases the potential of Alpha\textsc{Flow}-Lit in enabling faster and more scalable generation of protein ensembles.
\end{abstract}

\section{Introduction}

Exploring conformational landscapes is essential to capture the dynamic nature of protein structures, offering insights into their flexibility, biological function, and interactions. 
Traditionally, ensembles of conformational changes are collected through molecular dynamics (MD) simulations \cite{karplus2002molecular}.
While MD-base methods adhere to physical laws and will theoretically explore the entire landscape, they are time- and resource-intensive.
To expedite this process, some methods focus on increasing the diversity of AlphaFold \cite{jumper2021highly}, which is a powerful deep learning model for crystal structure prediction but falls short in accounting for conformational divergence.
Specifically, these methods sample different multiple sequence alignments (MSAs) as input \cite{wayment2024predicting} or enable the dropout function in AlphaFold \cite{wallner2023afsample} during the inference process.
Despite these inference interventions indeed bring some diversity to AlphaFold, they fall significantly short of generating enough conformational heterogeneity to thoroughly explore the protein landscape.

\begin{figure*}[htbp]
    \centering
    \includegraphics[width=\textwidth]{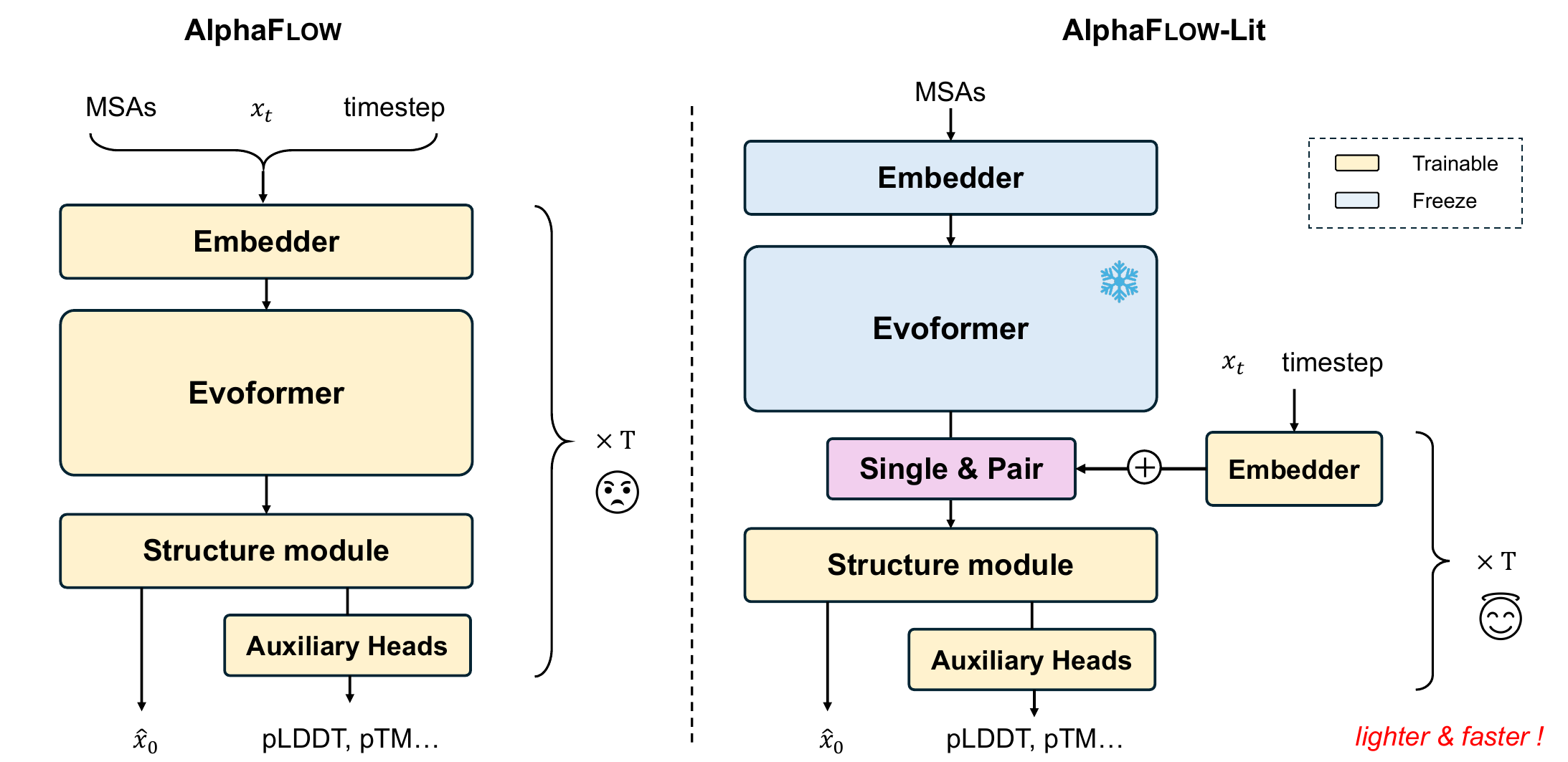}
    \caption{
        Model architecture of \textit{sequence-conditioned} Alpha\textsc{Flow} (\textit{left}) and \textit{feature-conditioned} Alpha\textsc{Flow}-Lit (\textit{right}).
        $T$: Denoising steps; $x_t$: Noisy structure; $\tilde{x}_0$: Predicted structure.
    }
    \label{fig:arch}
\end{figure*}

Recently, \cite{jing2024alphafold} harnessed the power of generative methods \textit{flow matching}, and integrated this framework into AlphaFold, called Alpha\textsc{Flow}.
To be concrete, it treats AlphaFold as a powerful \textit{sequence-conditioned} denoising model, which receives the noisy structures as templates and samples the protein ensembles from harmonic prior under a flow field.
Alpha\textsc{Flow} inherits the weights of AlphaFold, and was trained on general PDB then fine-tuned on different protein MD trajectories as a regression model, using loss functions similar to those in the original AlphaFold.
Due to these enhancements, Alpha\textsc{Flow} is much more flexible and diverse than the aforementioned inference intervention methods.
It is the first method to ingeniously combine both the advantages of accurate structure prediction and the generative capability for conformation sampling.

\begin{figure*}[htbp]
    \centering
    \includegraphics[width=\textwidth]{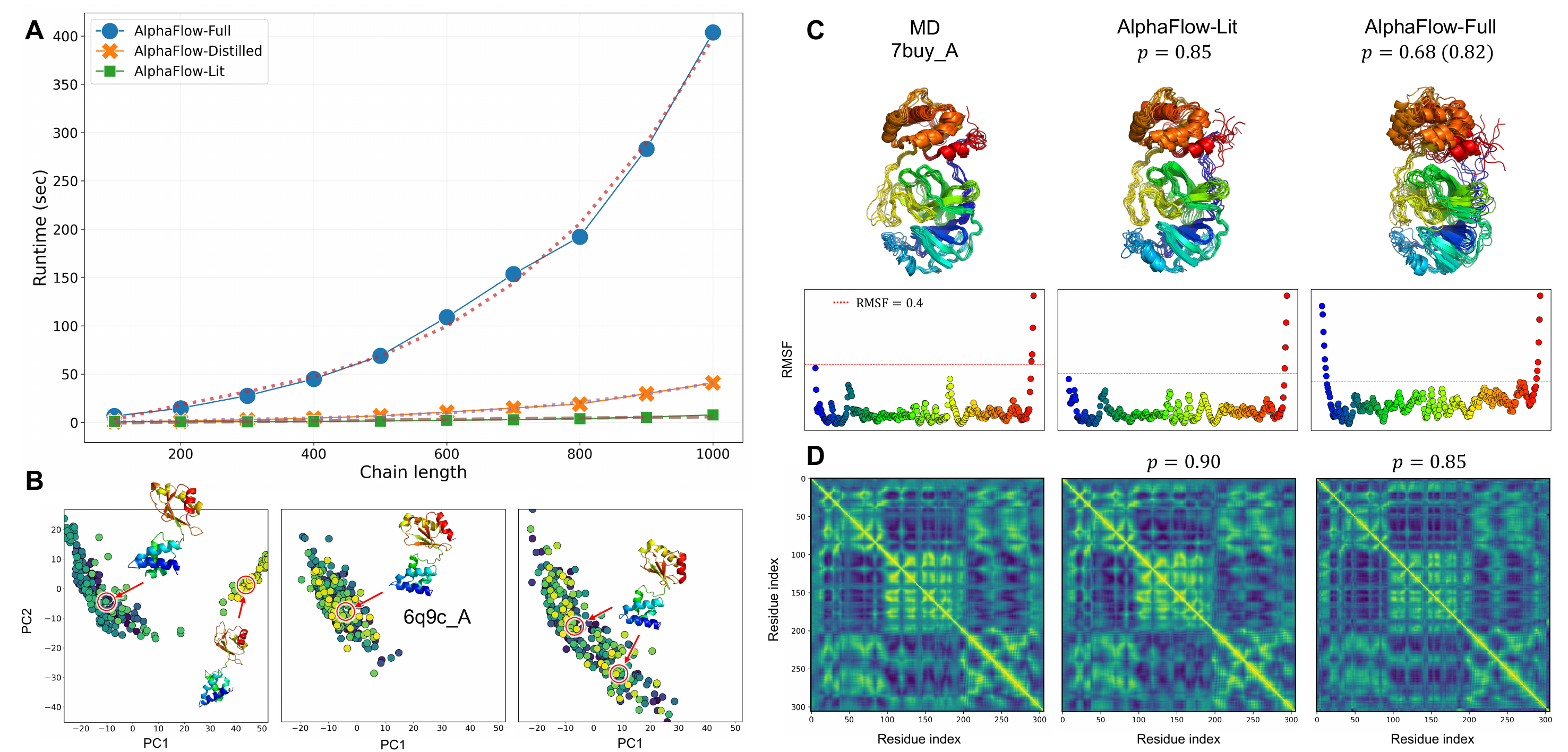}
    \caption{
        \textbf{Visualization of MD evaluation from MD, Alpha\textsc{Flow}-Lit and Alpha\textsc{Flow}.}
        (\textbf{A}) Runtime comparison corresponding to the sequence length and their fitted curves.
        (\textbf{B}) Principal components analysis (PCA) for \texttt{6q9c\_A} ensembles.
        The representative structures are pointed out.
        (\textbf{C, D}) Ensembles of PDB ID \texttt{7buy\_A} with C$\alpha$ RMSF by residue index shown in insets, and their Dynamic cross-correlation matrix (DCCM).
    }
    \label{fig:analysis}
\end{figure*}

However, limitations persist in sampling consumption.
As shown in Fig.~\ref{fig:arch}, since Alpha\textsc{Flow} is trained by fully fine-tuning AlphaFold,
generating the final structure $\hat{x}_0$ requires $T$ denoising steps, which means running $T$ times AlphaFold with additional embedders.
Although flow matching method is relatively faster compared with other diffusion methods, as shown in Fig.~\ref{fig:analysis}(\textbf{A}), Alpha\textsc{Flow} showcases cubic growth with the chain length, which leads to unacceptable time consumption and hinders its application for generating larger set of protein ensembles.
While Alpha\textsc{Flow} adopts diffusion distillation to reduce the generative process to a single forward pass, this approach compromises the level of the sampling performance.

To address this issue, we propose a \textit{feature-conditioned} generative model called Alpha\textsc{Flow}-Lit, which can be treated as an efficient and \textit{lighter} version of Alpha\textsc{Flow}.
As demonstrated in AlphaMissense \cite{cheng2023accurate}, features derived from MSAs encoder could be further utilized for variants classification.
The lastest AlphaFold3 also employs these features to train a non-equivariant denoiser \cite{abramson2024accurate} for the adaption of multi-modality such as nucleic acids, small molecules, ions, and modified residues.
Also, as shown in \cite{jing2024alphafold}, the MSAs have relatively minor impact on structural diversity compared with the flow matching framework.
Inspired by these findings, as illustrated in Fig.~\ref{fig:arch}, Alpha\textsc{Flow}-Lit retains the original AlphaFold embedder and Evoformer blocks in a frozen state and is directly conditioned on computed single and pair features.
Given that the remaining structure module and auxiliary heads are significantly lighter than the Evoformer block, compared to Alpha\textsc{Flow}, Alpha\textsc{Flow}-Lit can achieve a faster sampling process (around 47 times speedup) with the same number of denoising steps.
When directly trained on ATLAS dataset \cite{vander2024atlas} of protein MD trajectories, Alpha\textsc{Flow}-Lit performs on-par with Alpha\textsc{Flow} and surpasses its distilled version.
We also provide the analysis of the generated MD ensembles from Alpha\textsc{Flow}-Lit, including protein dynamics analysis, local arrangements within residue, and long-range correlations among residues to illustrate the diverse capabilities of Alpha\textsc{Flow}-Lit.

\section{Preliminary}

In this section, we briefly introduce the flow matching framework and some details of Alpha\textsc{Flow}.
\paragraph{Flow matching}
The flow matching framework begins with the continuous normalizing flow (CNF) $\psi_t$, defined as the solution of an ordinary differential equation (ODE) governed by a time-dependent vector field $u_t: \frac{d}{dt}\psi_{t} = u_t(\psi_t)$.
Let $x$ be a data point on a specific manifold, the CNF has an initial condition $\psi_0(x)=x$.
Given two distributions $p_0$ and $p_1$, we can define a probability path $p_t$ as their interpolation, which can be viewed as paths generated by $u_t$.
To effectively learn the CNF, we make $u_t$ tractable and by adopting the conditional probability path $p_t(x|x_1)$, which samples $x_0$ from prior distribution $p_0$ and interpolating it linearly with the data point $x_1$:
\begin{equation}
    x_t = (1 - t) \cdot x_0 + t \cdot x_1
\end{equation}
with the corresponding vector field:
\begin{equation}
    u_t(x_t|x_1) = (x_1 - x_t) / (1 - t)
\end{equation}
This method is referred as conditional flow matching (CFM).
We employ a neural network $v_t^{\theta}$ to learn the vector field.
The objective of CFM can be written as:
\begin{equation}
    \mathcal{L}_{\operatorname{CFM}} = \mathbb{E}_{t, p(x_0), p(x_1)}\left \|v_t^{\theta}(x_t) - u_t(x_t|x_1)\right \|^2
\end{equation}
\paragraph{Alpha\textsc{Flow}}
To integrate the AlphaFold, which is a regression model that directly outputs $x_1$, into the flow matching framework, Alpha\textsc{Flow} reparameterizes the neural vector field as:
\begin{equation}
    v_t^{\theta}(x_t) = (\operatorname{AlphaFold}(x_t) - x_t) / (1-t)
\end{equation}
It allows the objective to be rewritten as learning the expectation of $x_1$. 
Consequently, Alpha\textsc{Flow} can employ the similar regression loss function (e.g. FAPE) to optimize the neural network.
Alpha\textsc{Flow} introduces two key innovations:
(a) It employs the 3D coordinates of its $\beta$-carbons ($\alpha$-carbon for glycine) to describe the noisy structure and the prior distribution is defined over the $\beta$-carbons coordinates as a harmonic prior \cite{jing2023eigenfold};
(b) Alpha\textsc{Flow} treats $x_t$ as features (similar to templates), and the denoising process does not directly apply to the spatial domain as in prevailing SE(3) generative models \cite{yim2023fast, bose2023se, li2024f}. 
Instead, it starts from the identity rigids, which is the same as AlphaFold.
These contributions make Alpha\textsc{Flow} as a new paradigm for utilizing AlphaFold within different frameworks or applications.

\section{Method}

Alpha\textsc{Flow}-Lit follows the same ideas of Alpha\textsc{Flow} but introduces some modifications to the input pipeline.
Alpha\textsc{Flow}-Lit is a \textit{feature-conditioned} generative model, that is to say, it is conditioned on the single and pair features after the Evoformer blocks to generate diverse conformations.
As illustrated in Fig.~\ref{fig:arch},  the AlphaFold embedders (including the original input embedder, recycling embedder, extra MSA embedder, and extra MSA stack) and Evoformer are kept frozen.
The input embedding module for noisy structures $x_t$ and timesteps is similar to that of Alpha\textsc{Flow} but with little modifications.
The single and pair output of input embedding module are derived from the torsion angles (if designated) and contact map of the noisy structure, respectively.
These outputs are followed by a Linear layer initialized with zeros before summation with the features after Evoformer blocks.
This is similar to the zero convolution in ControlNet \cite{zhang2023adding}, designed to minimally disrupt the pretrained weights at the outset.
The detailed algorithm for input embedding module is described in Appendix A Algorithm \ref{alg:embedding}.
It is worth noting that the torsion angles in AlphaFold are represented in 8 rigids groups with $\sin-\cos$ formats, indicating rotation towards the coordinates of the former group. 
As a result, these angles are invariant to rigid transformations, eliminating the need to rotate the predicted structure after RMSD alignment.
The training procedure for Alpha\textsc{Flow}-Lit is the same as for Alpha\textsc{Flow}, and the inference procedure is provided in Algorithm \ref{alg:inference}.
We keep the Algorithm notations same as \cite{jing2024alphafold}.
The acceleration primarily results from the pre-computation of single and pair features. 
In contrast to Alpha\textsc{Flow}, Alpha\textsc{Flow}-Lit does not necessitate running Evoformer blocks at each denoising step but only conducts this process once at the inception.
Under this circumstances, the denoising network in Alpha\textsc{Flow}-Lit is a lightweight structure module that is conditioned on single and pair features rather than MSAs.
It distinguishes Alpha\textsc{Flow}-Lit as feature-conditioned and contributes to its efficiency and speed in generating protein emsembles.

\begin{table*}[htbp]
\centering
\caption{\textbf{Evaluation on MD ensembles.} We compare the predicted ensemble from Alpha\textsc{Flow}-Lit and Alpha\textsc{Flow} with the ground truth MD ensemble according to various metrics.
For Pairwise RMSD and Per-target RMSF, the ground truth values (from the MD ensembles) are in parenthesis.
Following \cite{jing2024alphafold}, the median across the 82 test ensembles is reported. $p$: Pearson correlation; $JSD$: Jensen-Shannon divergence.}
\label{tab:md}
\resizebox{\textwidth}{!}{%
\begin{tabular}{llccc}
\toprule
                                                  &                            & Alpha\textsc{Flow}-Full & Alpha\textsc{Flow}-Lit & Alpha\textsc{Flow}-Distilled \\ \midrule
\multirow{4}{*}{Protein dynamics}             & Pairwise RMSD ($=2.90$)     & 2.89           & 2.43          & 1.94                \\
                                                 & Pairwise RMSD $p$ $\uparrow$            & 0.49           & 0.58          & 0.49                \\
                                                 & PCA $C_{\alpha}$ coordinates $JSD$ $\downarrow$       & 0.43           & 0.46          & 0.51                \\
                                                  & PCA $C_{\alpha}$ pairwise distance $JSD$ $\downarrow$ & 0.48           & 0.52          & 0.56                \\
                                                  \midrule
\multirow{4}{*}{Local arrangements within residues}   
                                                  & Per-target RMSF ($=1.94$)   & 1.88           & 1.65          & 1.34                \\
                                                  & Per-target RMSF $p$ $\uparrow$         & 0.75           & 0.77          & 0.71                \\
                                                  & Stable contacts $JSD$ $\downarrow$                  & 0.84           & 0.83          & 0.79                \\
                                                  & Dihedral distributions $JSD$ $\downarrow$                 & 0.47           & 0.51          & 0.57                \\ \midrule
\multirow{1}{*}{Long-range correlations among residues} & DCCM $p$ $\uparrow$                     & 0.78           & 0.78          & 0.74                \\
                                                   \bottomrule
\end{tabular}%
}
\end{table*}

\begin{algorithm}[H]
\caption{\textsc{Inference}}\label{alg:inference}
\begin{algorithmic}
\STATE \textbf{Input:} Sequence and MSA $(A, M)$
\STATE \textbf{Output:} Sampled all-atom structure $\hat S$
\STATE Sample $\bfx_0 \sim \HarmonicPrior(\length(A))$ 
\STATE $\bff^{\text{tor}}_0 = \emptyset$
\STATE $\bfs_{i}^{\text{evo}}, \bfz_{ij}^{\text{evo}} \gets \Evoformer(\AlphaFoldEmbedder(A, M))$
\FOR{$n \gets 0$ to $N - 1$}
\STATE    Let $t \gets n / N$ and $s \gets t + 1/N$ \;
\STATE    $\bfs_{i}, \bfz_{ij} \gets \InputEmbedder(\bfx_t, \bff^{\text{tor}}_t, t)$
\STATE    $\bfs_{i}, \bfz_{ij} \pluseq \bfs_{i}^{\text{evo}}, \bfz_{ij}^{\text{evo}}$
\STATE    Predict $\hat S \gets \StructureModule(\bfs_{i}, \bfz_{ij}, A)$ \;
\IF{$n = N-1$}
\STATE \textbf{return} $\hat S$ 
\ENDIF
\STATE    Extract $\hat \bfx_1 \gets \BetaCarbons(\hat S)$ \;
\STATE    Align $\bfx_t \gets \RMSDAlign(\bfx_t, \hat \bfx_1)$ \;
\STATE    Interpolate $\bfx_s \gets \frac{s - t}{1-t} \cdot \hat \bfx_1 + \frac{1-s}{1-t} \cdot \bfx_t$ \;
\IF{embed\_angles}
\STATE    Extract $\bff^{\text{tor}}_t \gets \TorsionAngle(\hat S)$ \;
\ELSE 
\STATE    $\bff^{\text{tor}}_t \gets \emptyset$ \;
\ENDIF
\ENDFOR
\end{algorithmic}
\end{algorithm}

\section{Experiments}

We directly train Alpha\textsc{Flow}-Lit on ALTAS MD trajectories \cite{vander2024atlas} without pretraining on the PDB.
Similar to Alpha\textsc{Flow}, we use 1265/39/82 ensembles for the training, validation, and test splits, respectively. 
All multiple sequence alignments (MSAs) are derived from OpenProteinSet \cite{ahdritz2024openfold}. 
For sequences not present in OpenProteinSet, we use MMseqs2 \cite{steinegger2017mmseqs2} to search the UniRef30 and ColabDB databases \cite{mirdita2022colabfold}.
The initial weight of Alpha\textsc{Flow}-Lit is from the AlphaFold's publicly available weights.
ALTAS provides three parallel trajectories with 10,001 frames for each protein. 
We subsample the trajectories with a stride of 100 frames to create the training set (300 frames in total). 
During training, we uniformly sample one frame at each step.
Since the Evoformer blocks are frozen, we set the weight of masked MSA loss to 0.

We generate 250 samples for 82 targets in the test set.
Alpha\textsc{Flow}-Full refers to Alpha\textsc{Flow} with 10 consecutive denoising steps.
Alpha\textsc{Flow}-Distilled denotes its distilled with a single forward denoising step.
Alpha\textsc{Flow}-Lit employs the full denoising steps.
The protein ensembles of Alpha\textsc{Flow}-Full and Alpha\textsc{Flow}-Distilled are downloaded from its public repository\footnote{\href{https://github.com/bjing2016/alphaflow}{https://github.com/bjing2016/alphaflow}}.
We assess the sampling runtime based on the protein length for these methods and approximate their consumption curve.
To evaluate the effectiveness of each method, we first investigate the protein dynamics, considering both the general dynamics indicated by the pairwise root-mean-square deviation (RMSD) and the essential dynamics uncovered through principal components analysis (PCA) \cite{amadei1993essential}. 
Furthermore, we assess the detailed capability of each method at the residue resolution of protein dynamics by systematic comparisons of local arrangements within residues and motional correlations among residues. The results are presented in Table~\ref{tab:md} and Fig.~\ref{fig:analysis}.

\paragraph{Runtime comparison}
In Figure~\ref{fig:analysis}, we depict the relation between runtime of sampling and sequence length ranging from 100AA to 1,000AA in increments of 100. 
Alpha\textsc{Flow}-Lit demonstrate superior scalability, maintaining consistently low runtime across increasing protein lengths.
Alpha\textsc{Flow} exhibit cubic growth in runtime, indicating its inefficiency for longer chains.
This inefficiency could be attributed to the cubic complexity of attention in the Evoformer block.
While Alpha\textsc{Flow}-Distilled performs better than Alpha\textsc{Flow}-Full due to its single forward inference, it still shows moderate increases in runtime as protein length grows. 
 In summary, Alpha\textsc{Flow}-Lit surpasses Alpha\textsc{Flow} by 6 to 51 times (47 times in average) and Alpha\textsc{Flow}-Distilled by 2 to 4 times (3.8 times in average), making it the most efficient configuration and highlighting its potential for generating a larger set of protein ensembles.

\paragraph{Protein dynamics analysis}
For each conformational ensemble, the general dynamics are quantified as the average $C_{\alpha}$-RMSD between any pair of conformations. Using this measurement, the Alpha\textsc{Flow}-Lit ensembles demonstrate the strongest Pearson correlation with the ground truth ensembles produced by classic MD, while maintaining a comparable level of diversity in the conformational ensembles relative to Alpha\textsc{Flow}-Full. In contrast, Alpha\textsc{Flow}-Distilled loosely match the general dynamics with the ground truth and does not achieve the same level of diversity. Also, we assess the essential dynamics of proteins by projecting the ensembles onto the first two principal components (PCs) derived from PCA. Two common featurization methods for proteins are utilized: aligned $C_{\alpha}$ absolute coordinates and pairwise $C_{\alpha}$ internal distances. The differences in the distributions are quantified using the mean Jensen-Shannon divergence (JSD) for each PC between the predicted and true ensembles. Likewise, Alpha\textsc{Flow}-Lit exhibits essential dynamics distributions that are comparable to those of Alpha\textsc{Flow}-Full, surpassing the performance of Alpha\textsc{Flow}-Distilled. 
We visualize an example \texttt{6q9c\_A} in Fig.~\ref{fig:analysis}(\textbf{B}) and extract the representative structures. 
Both Alpha\textsc{Flow}-Full and Alpha\textsc{Flow}-Lit highly align with one of the principal component distributions of molecular dynamics (MD)
However, they do not capture the other distribution, which, although relatively minor, is still significant.

\paragraph{Local arrangements analysis}
Allostery, which has been coined the second secret of life after genetic codes, is a fundamental mechanism underlying most protein dynamics \cite{fenton2008allostery}. To further evaluate the practical effects of each method in identifying residues that undergo conformational changes in the local environment—changes that are likely critical for allostery—we performed a multifaceted analysis at the residue resolution, including thermally averaged flexibility, residual contact probabilities, and key dihedral angles distribution. In terms of thermally averaged flexibility, we calculate the root-mean-square fluctuation (RMSF) at the residue level, represented by $C_{\alpha}$. Alpha\textsc{Flow}-Lit achieves a strong Pearson correlation of 0.77 between the predicted and actual ensembles within a target, whereas Alpha\textsc{Flow}-Distilled only exhibits a moderate Pearson correlation of 0.71. 
We visualize the RMSF of \texttt{7buy\_A} in Fig.~\ref{fig:analysis}(\textbf{C}). 
We observe that the decrease in Pearson correlation of Alpha\textsc{Flow} is due to the high diversity of the structure's end point. 
If we exclude the first 5 residues and recompute the RMSF Pearson correlation (values in parentheses), both Alpha\textsc{Flow} and Alpha\textsc{Flow}-Lit will yield identical results.
In addition, contact probability analysis is utilized to elucidate the conformational rearrangements concerning the relative positions and orientations of structural motifs. Meanwhile, their internal conformational displacements are more accurately represented through variations in key dihedral angles. For contact probability analysis, a stable contact is defined as $C_{\alpha}$ pairs that maintain contact (with a threshold of 7\text{ \AA}) in over 85\% of the conformational ensembles. The Jaccard similarity (JS) between the contact residue pairs is calculated between the predicted and the ground truth sets. Regarding key dihedral angles, in addition to the backbone phi and psi angles, the chi1 values are also included, as they may indicate significant side chain reorientations relevant to the formation of key polar or non-polar interactions. The results show that contact and dihedral distributions exhibit moderate consistency between the actual and predicted ensembles generated by Alpha\textsc{Flow}-Full and Alpha\textsc{Flow}-Lit, unlike the larger inconsistencies observed with Alpha\textsc{Flow}-Distilled. 

\paragraph{Long-range correlations analysis}
Finally, we analyze the motional correlations similarity among residues by calculating the dynamic cross-correlation map (DCCM) using the conformational ensembles generated by different methods \cite{hunenberger1995fluctuation}. Such correlations could reveal pivot residues that mediate long-range allosteric coupling \cite{mcclendon2009quantifying}. Alpha\textsc{Flow}-Lit shows higher similarity in these matrices compared to Alpha\textsc{Flow}-Distilled, underscoring its superior ability to capture couplings even among long-range residues.
We visualize the DCCM of \texttt{7buy\_A} in Fig.~\ref{fig:analysis}(\textbf{D}).

\section{Conclusion}

We propose Alpha\textsc{Flow}-Lit, an improved version of  Alpha\textsc{Flow} for efficient protein ensembles generation.
Compared with Alpha\textsc{Flow}, Alpha\textsc{Flow}-Lit is a feature-conditioned generative model that eliminates the heavy reliance on MSAs encoding blocks and utilizes computed features to produce a diverse range of conformations. 
By directly training on ATLAS,  Alpha\textsc{Flow}-Lit performs on-par with Alpha\textsc{Flow} while outperforming its distilled version, all while achieving a substantial acceleration in sampling speed of around 47 times.
In addition, we conduct a thorough analysis of protein dynamics, local arrangements, and long-range coupling within the generated ensembles.
The advantages of Alpha\textsc{Flow}-Lit make it capable of generating a larger set of protein ensembles, enabling us to more effectively explore the protein landscape using deep learning techniques.

\paragraph{Limitation and future work}
As illustrated in the Experiment section, Alpha\textsc{Flow}-Lit exhibits less diversity compared to Alpha\textsc{Flow}-Full, likely due to the absence of pretraining on the PDB or insufficient training on MD trajectories.
This will be addressed in future work. Additionally, in the PCA analysis of example \texttt{6q9c\_A}, both models fail to capture the additional conformation present in the ground truth MD distribution. Enhancing their capability to capture such nuances will be a focus of our future research.

\section*{Acknowledgement}

This paper is supported by National Key Research and Development Program of China (2023YFF1205103), National Natural Science Foundation of China (81925034), National Natural Science Foundation of China (62088102) and a grant from the Hong Kong Innovation and Technology Fund (Project No. ITS/241/21).
We thank the open-source codebases from OpenFold Team and Bowen Jing (Alpha\textsc{Flow}).
We thank Dr. Le Zhuo for valuable discussions.

\clearpage

\bibliography{ref}
\bibliographystyle{icml2024}

\newpage
\appendix
\onecolumn
\section{Method Details.}

We highlight the difference with Alpha\textsc{Flow} in yellow. 
$\operatorname{init}=\operatorname{`final'}$ indicates the weight and bias (if has) of $\Linear$ is initialized with 0.
Other notations are kept same as \cite{jumper2021highly} and \cite{jing2024alphafold}.
\begin{algorithm}[H]
\caption{\textsc{InputEmbedding}}\label{alg:embedding}
\begin{algorithmic}
\STATE \textbf{Input:} {Beta carbon coordinates $\bfx \in \mathbb{R}^{N \times 3}$, Torsion angles $\bff^{\text{tor}}$, time $t\in[0,1]$}
\STATE \textbf{Output:} {Input pair embedding $\bfz \in \mathbb{R}^{N \times N \times 64}$}
\STATE $\bfz_{ij} \gets \lVert \bfx_i - \bfx_j \rVert $ 
\STATE $\bfz_{ij} \gets \Bin(\bfz_{ij}, {\min}=3.25 \text{ \AA}, {\max}= 50.75 \text{ \AA},  N_\text{bins}=39)$ 
\STATE $\bfz_{ij} \gets \Linear( \mathcolorbox{yellow}{\Concat ( \OneHot(\bfz_{ij}), \bff_{ij}^{\operatorname{mask\_2d}})})$ 
\FOR{$l \gets 1$ to $N_\text{blocks} = 4$}
\STATE    $\{\bfz\}_{ij} \pluseq \TriangleAttentionStartingNode({\bfz_{ij}}, c = 64, N_\text{head} = 4)$   
\STATE    $\{\bfz\}_{ij} \pluseq \TriangleAttentionEndingNode({\bfz_{ij}}, c = 64, N_\text{head} = 4))$ 
\STATE    $\{\bfz\}_{ij} \pluseq \TriangleMultiplicationOutgoing({\bfz_{ij}}, c = 64)$   
\STATE    $\{\bfz\}_{ij} \pluseq \TriangleMultiplicationIncoming({\bfz_{ij}}, c = 64)$ 
\STATE    $\{\bfz\}_{ij} \pluseq \PairTransition({\bfz_{ij}}, n = 2)$ 
\ENDFOR
\STATE  $\bfz_{ij} \gets \Linear(\bfz_{ij}, \mathcolorbox{yellow}{\operatorname{init}=\operatorname{`final'}})$ 
\STATE $\bfz_{ij} \pluseq \Linear
(\GaussianFourierEmbedding(t, \mathcolorbox{yellow}{d=128}), \mathcolorbox{yellow}{\operatorname{init}=\operatorname{`final'}})$
\IF{embed\_angles and $\bff^{\text{tor}} \neq \emptyset$}
\STATE $\mathcolorbox{yellow}{\bfs_{i} \gets \Linear(\Concat( \OneHot(\bff^{\text{tor}}_i, \bff^{\text{mask\_tor}}) ))}$
\STATE $\mathcolorbox{yellow}{\bfs_{i} \gets \Linear(\bfs_{i}, \operatorname{init}=\operatorname{`final'})}$
\ELSE 
\STATE $\bfs_i \gets \mathbf{0}$ \;
\ENDIF
\end{algorithmic}
\end{algorithm}

\newpage
\section{Runtime comparison}

\begin{table}[H]
\centering
\caption{Sampling runtime across Alpha\textsc{Flow},  Alpha\textsc{Flow}-Lit and  Alpha\textsc{Flow}-Distilled in Fig.~\ref{fig:analysis}(\textbf{A}). Proteins are selected from ATLAS. All methods are conducted on a single A100 GPU. Runtime values are reported in second.}
\resizebox{.8\textwidth}{!}{%
\begin{tabular}{lcccc}
\toprule
PDB ID  & Seq. length & Alpha\textsc{Flow}-Full & Alpha\textsc{Flow}-Distilled & Alpha\textsc{Flow}-Lit \\ \midrule
5h6x\_A & 100        & 6.63                & 0.86                     & 0.76               \\
2q9r\_A & 200        & 14.75               & 1.48                     & 0.81               \\
2v4b\_B & 300        & 27.74               & 2.76                     & 0.85               \\
1ru4\_A & 400        & 44.98               & 4.46                     & 1.10               \\
2d5b\_A & 500        & 68.97               & 6.82                     & 1.50               \\
6zsl\_B & 603        & 108.96              & 10.75                    & 2.20               \\
6lrd\_A & 705        & 153.57              & 14.93                    & 3.00               \\
4ys0\_A & 824        & 192.23              & 19.06                    & 3.94               \\
3nci\_A & 903        & 283.16              & 29.83                    & 5.44               \\
1gte\_D & 1025       & 403.68              & 40.97                    & 7.89               \\ \bottomrule
\end{tabular}%
}
\end{table}


\end{document}